\PassOptionsToPackage{table,xcdraw}{xcolor}
\documentclass[10pt,twocolumn,letterpaper]{article}

\usepackage{cvpr}              
\usepackage{multirow}
\usepackage{graphicx}
\usepackage{pifont}
\usepackage[normalem]{ulem}
\usepackage{caption}
\usepackage{float}
\usepackage{amsmath}
\useunder{\uline}{\ul}{}
\usepackage[table,xcdraw]{xcolor}
\definecolor{cvprblue}{rgb}{0.21,0.49,0.74}
\usepackage[pagebackref,breaklinks,colorlinks,allcolors=cvprblue]{hyperref}

\def\paperID{8175} 
\def\confName{CVPR}
\def\confYear{2025}

\title{MFogHub: Bridging Multi-Regional and Multi-Satellite Data for \\ Global Marine Fog Detection and Forecasting}
\author{
Mengqiu Xu$^{1,*}$, Kaixin Chen$^{1,*}$, Heng Guo$^1$, Yixiang Huang$^1$,\\
Ming Wu$^{1,\dagger}$, Zhenwei Shi$^2$, Chuang Zhang$^{1,3,\dagger}$, Jun Guo$^1$ \\
$^1$Beijing University of Posts and Telecommunications, Beijing, China \\
$^2$Beihang University, Beijing, China $^3$Beijing Wuzi University, Beijing, China  \\
{\tt\small \{xumengqiu,chenkaixin,guoheng,huangyixiang,wuming,zhangchuang,guojun\}@bupt.edu.cn} \\
{\tt\small shizhenwei@buaa.edu.cn }
}
\begin{document}


\twocolumn[{%
\maketitle
\vspace{-1cm}
\begin{figure}[H]
\hsize=\textwidth 
\centering
\includegraphics[width=\textwidth]{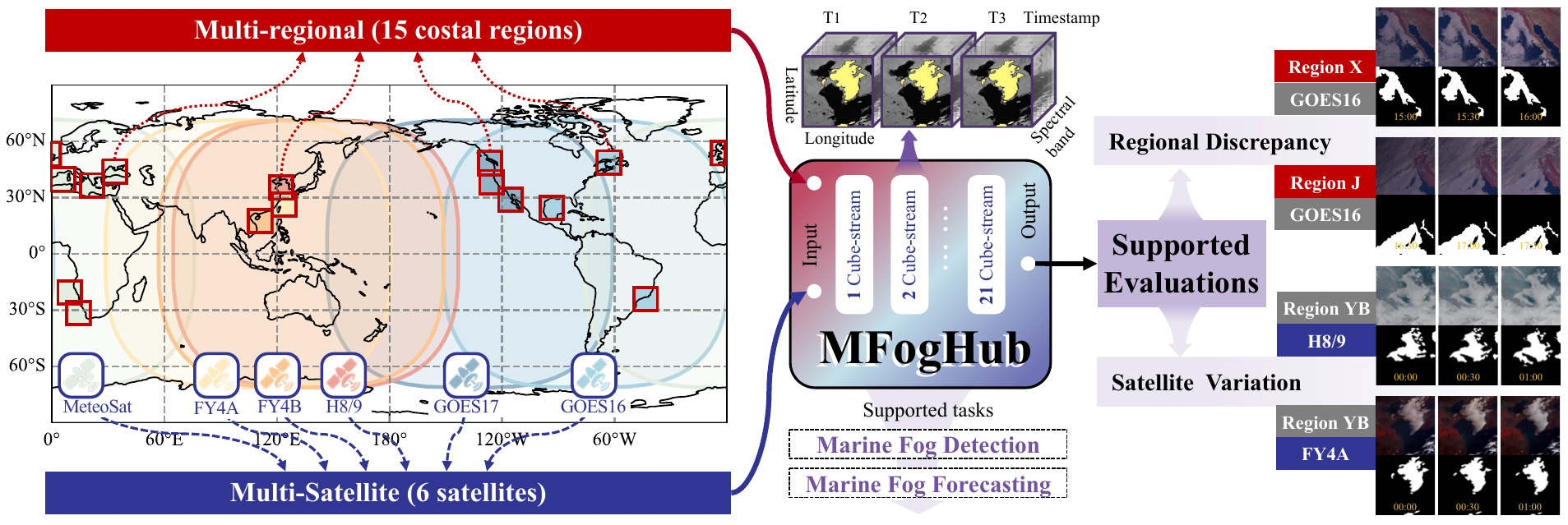}
\caption{\textbf{Overview of MFogHub}. 
\textbf{Right}: MFogHub collects data from 15 marine fog-prone regions worldwide, captured by 6 geostationary satellites. 
\textbf{Middle}: Data for each region-satellite pair is organized in a cube-stream structure with dimensions of ``timestamp-spectral band-latitude-longitude." MFogHub includes 21 cube-streams in total, each with corresponding masks, supporting both detection and forecasting tasks. 
\textbf{Left}: MFogHub enables unique evaluations of model generalization across multiple regions and satellite.}
\label{fig:teaser}
\end{figure}
}]

\let\thefootnote\relax\footnotetext{$\dagger$ Corresponding author, $^*$ Equal contribution.}
\begin{abstract}

Deep learning approaches for marine fog detection and forecasting have outperformed traditional methods, demonstrating significant scientific and practical importance. However, the limited availability of open-source datasets remains a major challenge. Existing datasets, often focused on a single region or satellite, restrict the ability to evaluate model performance across diverse conditions and hinder the exploration of intrinsic marine fog characteristics. 
To address these limitations, we introduce \textbf{MFogHub}, the first multi-regional and multi-satellite dataset to integrate annotated marine fog observations from 15 coastal fog-prone regions and six geostationary satellites, comprising over 68,000 high-resolution samples. 
By encompassing diverse regions and satellite perspectives, MFogHub facilitates rigorous evaluation of both detection and forecasting methods under varying conditions. 
Extensive experiments with 16 baseline models demonstrate that MFogHub can reveal generalization fluctuations due to regional and satellite discrepancy, while also serving as a valuable resource for the development of targeted and scalable fog prediction techniques. 
Through MFogHub, we aim to advance both the practical monitoring and scientific understanding of marine fog dynamics on a global scale. The dataset and code are at \href{https://github.com/kaka0910/MFogHub}{https://github.com/kaka0910/MFogHub}.
\end{abstract}

\vspace{-0.5cm}

\section{Introduction}
\label{sec:intro}

\begin{table*}[!ht]
\caption{Comparisons with existing marine fog detection and forecasting benchmarks: detailed overview of dataset information, geographic coverage, satellite types, event counts, supported tasks, and open accessibility. ($C_{sat}$ represents the number of spectral bands for different satellites, e.g. $C_{FY4A}=14$, $C_{H8/9}=16$)}
\label{tab:datasets_review}
\resizebox{\textwidth}{!}{%
\begin{tabular}{l|cccc|cc|cc|cc|cc|c}
\toprule

\multirow{2}{*}{\textbf{Datasets}} & \multicolumn{4}{c|}{\textbf{Information}} & \multicolumn{2}{c|}{\textbf{Regions}} & \multicolumn{2}{c|}{\textbf{Satellites}} & \multicolumn{2}{c|}{\textbf{Events}} & \multicolumn{2}{c|}{\textbf{Tasks}} &  \\

& \textbf{Num.} & \multicolumn{2}{c}{\textbf{Size $(H \times W)$~Spectral bands}} & \textbf{\small{Pixel-level labels}} & \textbf{Num.} & \textbf{Name} & \textbf{Num.} & \textbf{Types} & \textbf{w./w.o} & \textbf{Num.} & \textbf{Detection} & \textbf{Prediction} & \multirow{-2}{*}{\textbf{Open}} \\
\hline

\textbf{Nilo et al. (2018)}~\cite{meteo_seafog_2} & 72 & - & - & \ding{55} & 1 & Seas off Italy  & 1 & MeteoSat & \ding{51} & 9 & \ding{51} & \ding{55} & \textcolor[HTML]{d7191c}{\ding{55}} \\

\cellcolor[HTML]{F2F2F2}\textbf{Huang et al. (2021)}~\cite{seafog_dlinknet} & \cellcolor[HTML]{F2F2F2}201 & \cellcolor[HTML]{F2F2F2}( 1024,1024 ) & \cellcolor[HTML]{F2F2F2}16 & \cellcolor[HTML]{F2F2F2}\ding{51} & \cellcolor[HTML]{F2F2F2}1 & \cellcolor[HTML]{F2F2F2}Yellow and Bohai Sea & \cellcolor[HTML]{F2F2F2}1 & \cellcolor[HTML]{F2F2F2}H8/9 & \cellcolor[HTML]{F2F2F2}\ding{51} & \cellcolor[HTML]{F2F2F2}- & \cellcolor[HTML]{F2F2F2}\ding{51} & \cellcolor[HTML]{F2F2F2}\ding{55} & \cellcolor[HTML]{F2F2F2}\textcolor[HTML]{d7191c}{\ding{55}} \\

\textbf{Mahdavi et al. (2020)}~\cite{goes_seafog_1} & 53 & - & - & \ding{55} & 1 & The Grand Banks & 1 & GOES & \ding{51} & 26 & \ding{51} & \ding{55} & \textcolor[HTML]{d7191c}{\ding{55}} \\

\cellcolor[HTML]{F2F2F2}\textbf{Zhou et al. (2022)}~\cite{dual_branch} & \cellcolor[HTML]{F2F2F2}1,040 & \cellcolor[HTML]{F2F2F2}(512, 512) & \cellcolor[HTML]{F2F2F2}3 & \cellcolor[HTML]{F2F2F2}\ding{51} & \cellcolor[HTML]{F2F2F2}1 & \cellcolor[HTML]{F2F2F2}Yellow and Bohai Sea & \cellcolor[HTML]{F2F2F2}1 & \cellcolor[HTML]{F2F2F2}GOCI & \cellcolor[HTML]{F2F2F2}\ding{51} & \cellcolor[HTML]{F2F2F2}133 & \cellcolor[HTML]{F2F2F2}\ding{51} & \cellcolor[HTML]{F2F2F2}\ding{55} & \cellcolor[HTML]{F2F2F2}\textcolor[HTML]{1a9641}{\ding{51}} \\

\textbf{Tao et al. (2022)}~\cite{seafog_self} & 1,100 & ( 64, 64) & 16 & \ding{55} & 1 & Yellow and Bohai Sea & 1 & H8/9 & \ding{55} & - & \ding{51} & \ding{55} & \textcolor[HTML]{d7191c}{\ding{55}} \\

\cellcolor[HTML]{F2F2F2}\textbf{Huang et al. (2023)}~\cite{my_tgarss} & \cellcolor[HTML]{F2F2F2}4,291 & \cellcolor[HTML]{F2F2F2}( 1024, 1024) & \cellcolor[HTML]{F2F2F2}3 & \cellcolor[HTML]{F2F2F2}(356) & \cellcolor[HTML]{F2F2F2}1 & \cellcolor[HTML]{F2F2F2}Yellow and Bohai Sea & \cellcolor[HTML]{F2F2F2}1 & \cellcolor[HTML]{F2F2F2}H8/9 & \cellcolor[HTML]{F2F2F2}\ding{55} & \cellcolor[HTML]{F2F2F2}- & \cellcolor[HTML]{F2F2F2}\ding{51} & \cellcolor[HTML]{F2F2F2}\ding{55} & \cellcolor[HTML]{F2F2F2}\textcolor[HTML]{1a9641}{\ding{51}} \\

\textbf{Su et al. (2023)}~\cite{my_bmvc} & 1,808 & ( 1024, 1024 ) & 16 & \ding{51} & 1 & Yellow and Bohai Sea & 1 & H8/9 & \ding{55} & - & \ding{51} & \ding{55} & \textcolor[HTML]{d7191c}{\ding{55}} \\

\cellcolor[HTML]{F2F2F2}\textbf{Wu et al. (2024)}~\cite{S2DNet} & \cellcolor[HTML]{F2F2F2}193 & \cellcolor[HTML]{F2F2F2}( 2000, 1600 ) & \cellcolor[HTML]{F2F2F2}16 & \cellcolor[HTML]{F2F2F2}\ding{51} & \cellcolor[HTML]{F2F2F2}1 & \cellcolor[HTML]{F2F2F2}Yellow and Bohai Sea & \cellcolor[HTML]{F2F2F2}1 & \cellcolor[HTML]{F2F2F2}H8/9 & \cellcolor[HTML]{F2F2F2}\ding{55} & \cellcolor[HTML]{F2F2F2}23 & \cellcolor[HTML]{F2F2F2}\ding{51} & \cellcolor[HTML]{F2F2F2}\ding{55} & \cellcolor[HTML]{F2F2F2}\textcolor[HTML]{1a9641}{\ding{51}} \\

\textbf{Bari et al. (2023)}~\cite{meteo_seafog_1} & 12,744 & ( 661, 691) & 3 & \ding{55} & 1 & Seas around Morocco & 1 & MeteoSat & \ding{55} & - & \ding{55} & \ding{51} & \textcolor[HTML]{d7191c}{\ding{55}} \\

\cellcolor[HTML]{F2F2F2}\textbf{Mahdavi et al. (2024)}~\cite{fog_forecast} & \cellcolor[HTML]{F2F2F2}2,208 & \cellcolor[HTML]{F2F2F2}- & \cellcolor[HTML]{F2F2F2}- & \cellcolor[HTML]{F2F2F2}\ding{51} & \cellcolor[HTML]{F2F2F2}1 & \cellcolor[HTML]{F2F2F2}The Grand Banks & \cellcolor[HTML]{F2F2F2}1 & \cellcolor[HTML]{F2F2F2}GOES & \cellcolor[HTML]{F2F2F2}\ding{51} & \cellcolor[HTML]{F2F2F2}92 & \cellcolor[HTML]{F2F2F2}\ding{55} & \cellcolor[HTML]{F2F2F2}\ding{51} & \cellcolor[HTML]{F2F2F2}\textcolor[HTML]{d7191c}{\ding{55}} \\

\midrule
\textbf{MFogHub (Ours)} & 68,000 & (1024, 1024) & $C_{sat}$ & \ding{51} & 15 & \small{\begin{tabular}[c]{@{}c@{}}Yellow/Bohai Sea, East Sea, South Sae \\  Mediterranean,  North Sea, Gulf of \\ Alaska, California Current, etc.\end{tabular}} & 6 & \begin{tabular}[c]{@{}c@{}}FY4A, FY4B, \\ GOES16, GOES17, \\ H8/9, MeteoSat\end{tabular}  & \ding{51} & 693 & \ding{51} & \ding{51} & \textcolor[HTML]{1a9641}{\ding{51}} \\

\bottomrule
\end{tabular}%
}
\end{table*}

Marine fog is a complex and hazardous meteorological phenomenon that occurs in the lower atmosphere over oceanic regions~\cite{seafog_review_1,seafog_review_2}. During the marine fog formation process, horizontal visibility often drops below one kilometer~\cite{my_cja, huang}, causing significant disruptions to shipping, port operations, and coastal activities~\cite{meteo_seafog_1, cctv_seafog}. As such, research on marine fog detection and forecasting is of both scientific and practical importance.

In recent years, data-driven deep learning methods~\cite{seafog_dlinknet, seafog_scselinknet, eca_transunet, dual_branch} have increasingly outperformed traditional approaches~\cite{seafog_traditional_1, seafog_traditional_2, seafog_physics} in marine fog detection and forecasting. However, as shown in Table~\ref{tab:datasets_review}, the limited availability of open-source datasets in this area has led most studies~\cite{meteo_seafog_2, seafog_dlinknet, goes_seafog_1, seafog_self, my_tgarss, fog_forecast, dual_branch, S2DNet} to rely on region-specific or satellite-specific data collected in isolation. Consequently, early research has predominantly focused on local performance rather than the broader generalizability of models, potentially leading to inaccurate assessments of a method’s genuine capabilities. Furthermore, the reliance on single-region datasets has constrained exploration into the underlying mechanisms of marine fog formation and dissipation.


To address these issues, we introduce the \textbf{MFogHub} dataset—the first multi-regional, multi-satellite dataset for global marine fog detection and forecasting. MFogHub contains over 68,000 samples, as shown in Fig.~\ref{fig:teaser}, and spans 15 coastal fog-prone regions, consolidating 693 marine fog events. The dataset captures multi-band meteorological data from 6 geostationary satellites. Given the spatiotemporal nature of marine fog, we organize the data samples in a cube-stream structure with dimensions of `timestamp-spectral band-latitude-longitude' to facilitate usability. The minimum time interval is 30 minutes, with a spatial resolution of 1 km and a size of 1024 $\times$ 1024 pixels. Additionally, more than 11,600 samples are meticulously annotated at the pixel level by meteorological experts. We also provide visual statistical analysis to illustrate the sensitivity of marine fog across variations in region, satellite, and spectral band.



Sixteen standard detection and forecasting methods are selected to showcase the capabilities of MFogHub and establish benchmarks. Unlike single-region and single-satellite datasets, our proposed dataset enables a comprehensive and nuanced evaluation across diverse regions and satellites, revealing performance fluctuations across different methods due to regional discrepancies and satellite variations. Moreover, this dataset presents a new challenge in effectively leveraging multi-regional and multi-satellite data to uncover deeper characteristics of marine fog, thereby enhancing performance and advancing both detection and forecasting. Finally, we investigate the effects of spectral bands and the ratio of positive to negative samples on model performance, aiming to provide valuable insights for future research. 



\paragraph{Contributions} We introduce MFogHub, the first multi-regional (15 coastal regions) and multi-satellite (6 geostationary satellites) dataset for global marine fog detection and forecasting. This dataset is organized with a novel cube-stream structure, enabling efficient use in deep learning. Through experiments, we demonstrate that MFogHub supports comprehensive evaluation of model performance across regional and satellite dimensions, revealing insights into generalizability. Moreover, it provides a unique opportunity to explore multi-regional and multi-satellite data, facilitating the exploration of the complex dynamics of marine fog. We believe that MFogHub lays a crucial foundation for future research at the intersection of machine learning and meteorology.



\section{MFogHub}
\label{sec:m4fog}

\subsection{Multi-regional data collection}



\begin{figure}[ht]
\centering
\includegraphics[width=\linewidth]{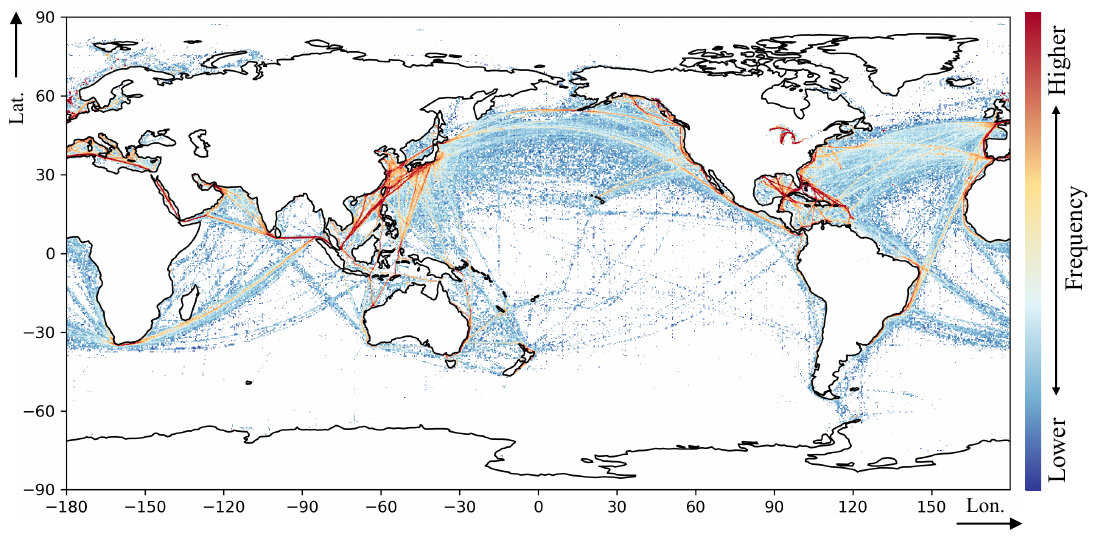}
\caption{Visualization of marine fog occurrence frequency based on 9.5 million ICOADS observations (2015-2024) for identifying costal fog-prone regions worldwide. }
\label{fig:merge}
\end{figure}

Marine fog is a global phenomenon that occurs in coastal areas~\cite{goes_seafog_1, fy3d_seafog_1, cctv_seafog, meteo_seafog_1} and extends to polar regions~\cite{arctic_fog, arctic_fog_1}, excluding tropical oceans, as long as the necessary temperature and humidity conditions are met. To better handle global marine fog, we extract almost 9.5 million records from all  International Comprehensive Ocean-Atmosphere Data Set (ICOADS)~\cite{icoads} data spanning 2015-2024, which clearly observe the presence or absence of marine fog globally. These records are accumulated into a global grid of 0.25° $\times$ 0.25° for frequency tallying, followed by statistical analysis and visualization, as illustrated in Fig.~\ref{fig:merge}. The areas in red on the map indicate higher frequencies of marine fog observations in the ICOADS data, while areas in blue represent lower frequencies. It is evident that maritime navigation is primarily concentrated along coastlines or established shipping routes, which is why this study focuses on coastal areas for marine fog detection and forecasting. Additionally, by applying a 12.8° sliding window across global regions and drawing on previous research~\cite{seafog_review_1,seafog_review_2}, we selected 15 coastal regions with high shipping traffic and frequent fog occurrences as the focus of this study. The detailed covering ranges and information of 15 marine regions are provided in Fig.~\ref{fig:teaser} and the Supplementary Materials. 

\subsection{Multi-satellite data collection}

Geostationary satellites, often referred to as the ``eyes in the sky," provide high spatial and temporal resolution data essential for marine fog detection and forecasting. In selecting satellites for this study, we consider two primary factors. First, we prioritiz coverage to capture a global perspective of marine fog. To achieve this, we incorporate data from China’s Fengyun~(FY)-4A and 4B, Europe’s MeteoSat, and the United States’ GOES-16 and GOES-17 satellites. The second factor is the ability to support multi-satellite studies within the same region. Therefore, we also included Japan’s Himawari~(H)-8/9 satellites, which cover regions similar to those of the Fengyun satellites. To improve observational detail and better capture diverse surface features, we utilized multi-spectral L1 data (excluding L1.5 data from MeteoSat) following the projection process. This data includes visible, near-infrared, and far-infrared channels. The primary differences among the data from these satellites lie in the number of spectral bands and their central wavelengths. Additionally, spatial resolution varies across channels, ranging from 500m to 4km. For consistency and comparability, all regional multi-band data were standardized to a spatial resolution of 1km ($\approx$0.0125°) after projection.


\subsection{Cube-stream structure} 



Marine fog is a dynamically evolving meteorological phenomenon that spans multiple stages, including formation, maintenance and dissipation. Continuous data samples are especially crucial for forecasting tasks. Thus, to better organize multi-regional, multi-satellite meteorological data for marine fog detection and forecasting, we propose a cube-stream data structure in the MFogHub building upon previous work~\cite{mesogeos}. 
As shown in middle part of Fig.~\ref{fig:teaser}, each region-satellite data is organized into a cube-stream along with the dimensions of ``timestamp-spectral band-latitude-longitude," represented as $\mathbb{R}^{T \times C \times H \times W}$, where $T, C, H, W$ denotes the number of timestamps, the number of spectral bands, height and width, respectively. Then all cube-stream can be further integrated across different regions and satellites to form the comprehensive MFogHub dataset.


To enable MFogHub to support validation across multiple dimensions, key attributes such as region, satellite, and time are customizable, allowing users to retrieve data that meets specific criteria and assemble custom sub-datasets. Such organization facilitates research into multi-regional and multi-satellite diversity and variation. Furthermore, each cube-stream data across spectral bands is spatio-temporally aligned, enabling flexible slicing operations across multiple dimensions for finer regional data extraction and spectral band sensitivity analysis.


\begin{figure}[ht]
  \centering
  \includegraphics[width=\linewidth]{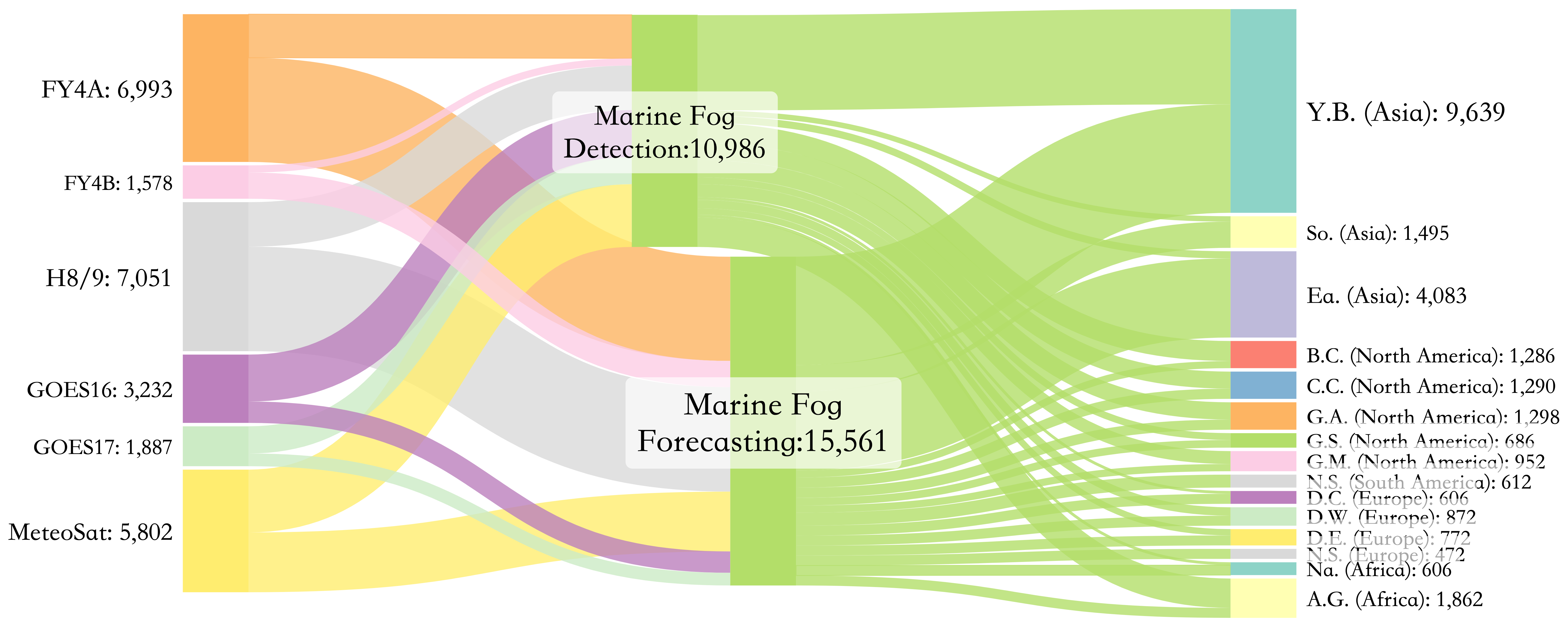}
  \caption{Illustration of data proportion and flow across various regions and satellites for marine fog detection and forecasting in the MFogHub dataset.}
  \label{fig:statistics}
\end{figure}


Finally, the MFogHub dataset integrates 21 data streams from 6 satellites and comprises 68,000 samples. It illustrates the distribution and flow of data across various satellites, regions, and corresponding tasks within the dataset as shown in the Fig.~\ref{fig:statistics}. 
Based on meteorological reports from various meteorological agencies, 11,600 of these samples are pixel-labeled to indicate the presence or absence of marine fog. Detailed annotation procedures are provided in the Supplementary Materials.

\section{Data analysis}
\label{sec:Statistics_Analysis}

\subsection{Regional discrepancy}

\begin{figure}[!t]
  \centering
  \includegraphics[width=\linewidth]{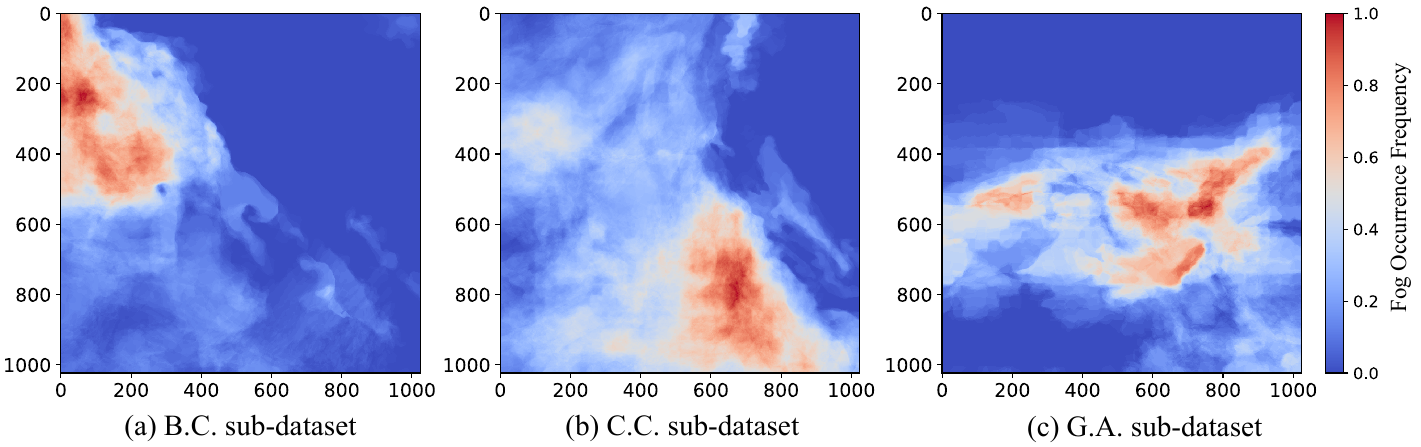}
  \caption{Spatial distribution and intensity variations across B.C., C.C. and G.A. sub-datasets using marine fog detection labels.}
  \label{fig:position_distribution}
\end{figure}

Marine fog distribution varies significantly across regions due to factors such as proximity to land, ocean currents, and other environmental conditions. Using marine fog data and labels from the MFogHub dataset for three regions—Baja California (B.C.), California Current (C.C.), and Gulf of Alaska (G.A.), observed by the GOES-16 satellite—this study demonstrates how the spatial distribution of marine fog occurrences differs across regions, as shown in Fig.~\ref{fig:position_distribution}. In the visualization, regions with redder hues correspond to higher frequencies of marine fog events.

It is clearly evident from the Fig.~\ref{fig:position_distribution} that each sub-dataset presents distinct spatial distribution patterns, with significant differences in intensity across regions. Additionally, the visualization reveals that the accumulation patterns of marine fog events vary between regions: the Baja California region exhibits more concentrated fog occurrences, while the other two regions show more dispersed patterns. These results indicate that the spatial distribution of marine fog differs notably across regions, posing a significant challenge to the model’s generalization ability.

\subsection{Satellite variation}

\begin{figure}[!t]
  \centering
  \includegraphics[width=\linewidth]{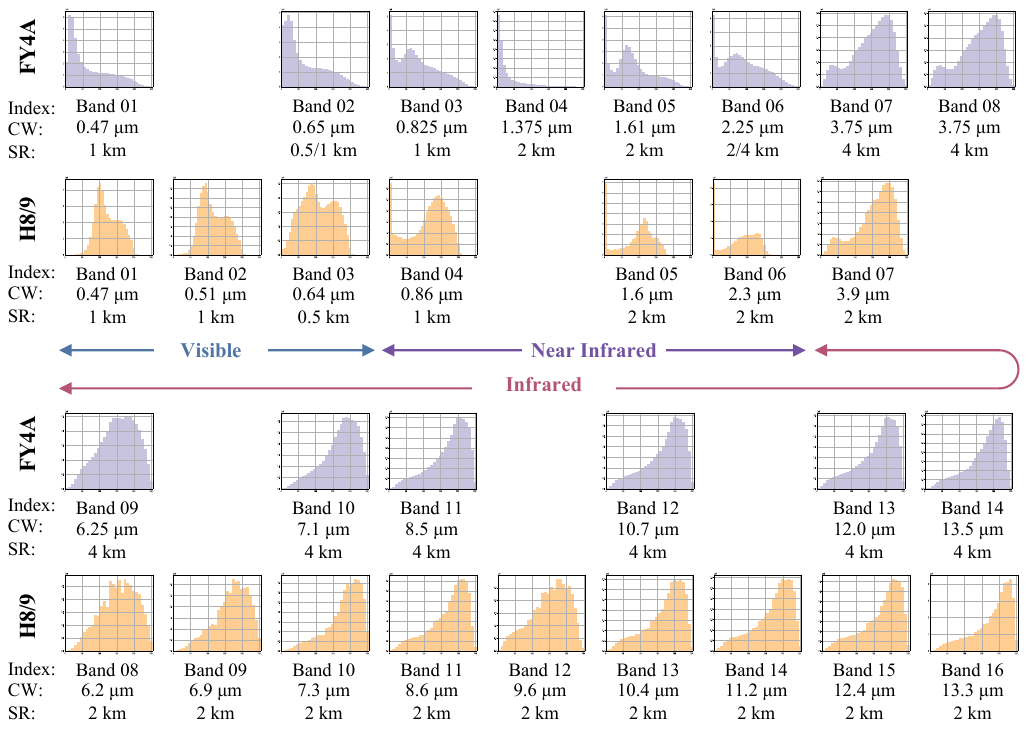}
  \caption{Visualization of pixel histogram distributions and band information (Index, Central Wavelength (CW), and Spatial Resolution (SR)) across multiple bands for FY4A and H8/9 satellites.}
  \label{fig:analysis_01}
\end{figure}

Geostationary satellites differ in imaging capabilities due to variations in spectral bands, spatial resolution, and sub-satellite points, leading to performance differences in models trained on data from different satellites for the same region. Therefore, using the adjacent FengYun-4A (FY4A) and Himawari-8/9 (H8/9) satellites as examples, we analyzed marine fog detection data from the Yellow and Bohai Seas during 2021. We computed pixel histogram distributions for each spectral band, with the visualized results and band information shown in Fig.~\ref{fig:analysis_01}. Here, purple and orange figures represent the 14 spectral bands of FY4A and 16 spectral bands of H8/9, respectively.


The most evident distinction is that H8/9 has an advantage in the infrared range, while FY4A has more coverage in the near-infrared. Some key spectral bands overlap between FY4A and H8/9, though with notable differences. For example, the band centered around 0.65 $\mu$m shows a bimodal distribution for H8/9 but a unimodal distribution for FY4A. Additionally, H8/9 has three visible bands, enhancing its capacity for true-color imaging. These differences underscore the complementary roles of FY4A and H8/9 in multi-spectral observation and suggest potential benefits from combining their data for marine fog detection.

\begin{figure*}[!t]
  \centering
  \includegraphics[width=\textwidth]{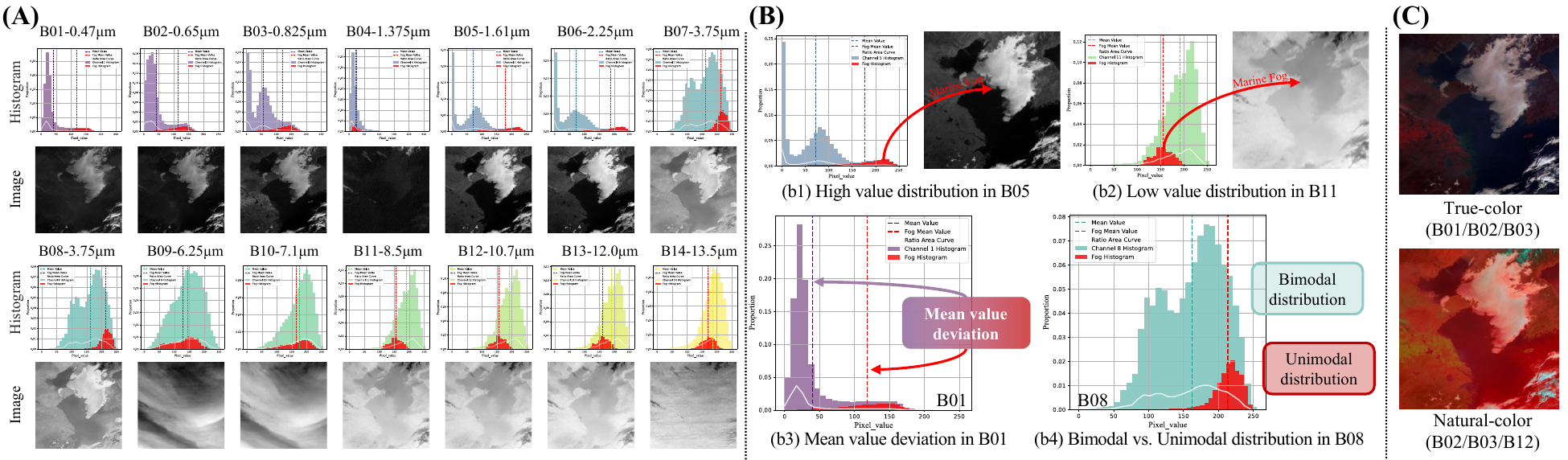}
  \caption{Sensitivity analysis of spectral band for one marine fog event captured by the FY-4A satellite on March 25, 2021. (A) Pixel histogram visualization comparison and single-band images between fog regions (red areas) and non-fog regions (colored areas). (B) Detailed description for characteristics and separability of marine fog area. (C) Synthesized true-color image and natural-color image.}
  \label{fig:analysis_02}
\end{figure*}


\subsection{Spectral band sensitivity of marine fog}

Due to the varied detection purposes of the satellite imaging instrument's multiple spectral bands, marine fog exhibits different sensitivities and image characteristics across spectral bands, which can be further reflected in the pixel value distributions. Figure.~\ref{fig:analysis_01} (A) presents a visualization of pixel histograms for each spectral band image from one marine fog event captured by the FY4A satellite, where the red and colored areas represent the pixel value distributions in fog and non-fog regions across different spectral band, respectively. It can be observed that the pixel distributions across different spectral bands show significant variation, closely linked to the central wavelength of each band.

Firstly, we observe that marine fog primarily exhibits two distinct characteristics across different spectral bands. Specifically, in the visible band (0.4-0.7 $\mu$m), as shown in Fig.~\ref{fig:analysis_02} (b1), water droplets within fog regions strongly scatter sunlight, causing marine fog areas to generally appear as high-brightness regions in the image. In contrast, in the water vapor or infrared bands, as illustrated in Fig.~\ref{fig:analysis_02} (b2), the water droplets in fog regions have weaker absorption of radiation, resulting in fog areas appearing as low-brightness regions. 

Additionally, the separability between marine fog (red areas) and non-fog regions (colored areas) varies across bands, with high separability evident in two specific scenarios. (1) As shown in Fig.~\ref{fig:analysis_02} (b3), the peak values of marine fog and non-fog regions are clearly separated into distinct pixel value ranges, with a significant mean difference that which are indicated by two dashed lines. (2) As illustrated in Fig.~\ref{fig:analysis_02} (b4), non-fog pixels exhibit a “bimodal” distribution, whereas fog pixels display a “unimodal” distribution. Finally, based on the previous analysis, we visualized true-color images synthesized from bands B01/B02/B03 and natural-color images synthesized from bands B02/B03/B12 as shown in Fig.~\ref{fig:analysis_02} (C), indicating that the combined use of multiple spectral bands may enhance model performance in distinguishing fog from non-fog regions.

\section{Experiments}
In this section, we evaluate the capability of MFogHub to assess the regional and satellite generalization of marine fog models, emphasizing its potential to support multi-region and multi-satellite research. Additionally, we investigate the impact of satellite spectral band selection and the positive-to-negative sample ratio on model performance, providing insights for future studies.


\subsection{Setup}
\paragraph{Marine fog detection} involves identifying and delineating marine fog regions from satellite data. Formally, the model is trained to map the input data $ \mathcal{X} \in \mathbb{R}^{C \times H \times W}$ to a binary mask $ \mathcal{M} \in \mathbb{R}^{ H \times W}$, where each pixel value $ \mathcal{M}_{i,j} $ indicates the presence or absence of marine fog at pixel location $(i,j)$. Here, $C$, $H$, and $W$ represent the number of spectral bands, and the latitude and longitude dimensions of a single data sample, respectively.

\paragraph{Marine fog forecasting} involves predicting future image sequences based on historical images. Formally, the model is trained to map a sequence of data samples $ \mathcal{X}^{t,T} \in \mathbb{R}^{T \times C \times H \times W}$, where $T$, $C$, $H$, and $W$ represent the temporal length, spectral bands, latitude, and longitude, respectively, to a future sequence $ \mathcal{Y}^{t+1,T'} \in \mathbb{R}^{T' \times C \times H \times W}$. 

\subsection{Implementation details}
\paragraph{Detection baselines and metrics.} Eight baseline models are selected for marine fog detection, including CNN-based architectures (Deeplabv3+~\cite{deeplabv3+}, UNet~\cite{unet}, UNet++~\cite{unetpp}), ViT-based architectures (ViT~\cite{vit}, DlinkViT~\cite{my_rs}), hybrid structures (Unetformer~\cite{unetformer}) and remote sensing methods (BANet~\cite{banet}, ABCNet~\cite{abcnet}). To evaluate their performance on the MFogHub dataset, we used the following metrics: Critical Success Index (CSI), recall, precision, Mean Accuracy (mAcc), and Mean Intersection over Union (mIoU).

\paragraph{Forecasting baselines and metrics.} Eight forecasting models are selected and categorized into two main types, following OpenSTL~\cite{openstl}. One category consists of recurrent-based methods, including ConvLSTM~\cite{convlstm}, PredRNN~\cite{predrnn}, MIM~\cite{mim}, and PhyDNet~\cite{phydnet}, while the other category includes recurrent-free methods such as SimVP v2~\cite{simvpv2}, Uniformer~\cite{uniformer}, VAN~\cite{van}, and TAU~\cite{tau}. In addition to standard performance evaluation metrics such as Mean Squared Error (MSE) and Mean Absolute Error (MAE), we also used Structural Similarity Index Measure (SSIM) and Peak Signal-to-Noise Ratio (PSNR)~\cite{metrics} to assess the forecasting models.

\begin{table*}[!t]
\caption{Quantitative results of marine fog detection baselines across three sub-regions (B.C., C.C. and G.A.) from the GOES satellite data.}
\label{tab:XJS_exps}
\resizebox{\linewidth}{!}{%
\begin{tabular}{l|ccccc|ccccc|ccccc}
\toprule
\multicolumn{1}{l|}{} & \multicolumn{5}{c|}{\textbf{B.C. sub-dataset from GOES}} & \multicolumn{5}{c|}{\textbf{C.C. sub-dataset from GOES}} & \multicolumn{5}{c}{\textbf{G.A. sub-dataset from GOES}} \\
\multicolumn{1}{l|}{} & \textbf{CSI$\uparrow$} & \textbf{Rec$\uparrow$} & \textbf{Pre$\uparrow$} & \textbf{mIoU$\uparrow$} & \textbf{mAcc$\uparrow$} & \textbf{CSI$\uparrow$} & \textbf{Rec$\uparrow$} & \textbf{Pre$\uparrow$} & \textbf{mIoU$\uparrow$} & \textbf{mAcc$\uparrow$} & \textbf{CSI$\uparrow$} & \textbf{Rec$\uparrow$} & \textbf{Pre$\uparrow$} & \textbf{mIoU$\uparrow$} & \textbf{mAcc$\uparrow$} \\

\midrule
\textbf{Deeplabv3p}~\cite{deeplabv3+} & 20.73 & 22.39 & 73.62 & 58.37 & 61.00 & 51.17 & 55.25 & \textbf{87.38} & 73.81 & 77.35 & 27.40 & 46.66 & 39.89 & 61.25 & 71.91 \\
\textbf{UNet}~\cite{unet} & 31.30 & 40.15 & 58.68 & 63.58 & 69.39 & 46.53 & 57.96 & 70.24 & 71.02 & 78.13 & 27.01 & 43.60 & 41.53 & 61.18 & 70.56 \\
\textbf{Unet++}~\cite{unetpp} & {\ul 43.58} & \textbf{72.20} & 52.36 & 69.56 & \textbf{84.52} & 50.52 & {\ul 69.73} & 64.70 & 72.94 & 83.55 & 23.79 & 47.71 & 32.18 & 58.86 & 71.82 \\
\textbf{ABCNet}~\cite{abcnet} & 21.68 & 24.22 & 67.36 & 58.80 & 61.83 & {\ul 61.97} & \textbf{71.46} & 82.35 & {\ul 79.50} & {\ul 85.20} & 32.53 & 37.46 & \textbf{71.21} & 64.73 & 68.42 \\
\textbf{BANet}~\cite{banet} & 37.74 & 40.67 & 83.97 & 67.30 & 70.15 & \textbf{63.03} & 71.42 & {\ul 84.28} & \textbf{80.09} & \textbf{85.25} & \textbf{43.22} & \textbf{72.89} & 51.50 & \textbf{69.69} & \textbf{85.05} \\
\textbf{Unetformer}~\cite{unetformer} & 42.54 & {\ul 56.05} & 63.84 & {\ul 69.48} & {\ul 77.26} & 49.76 & 62.94 & 70.38 & 72.73 & 80.55 & 34.16 & 51.96 & 49.93 & 65.09 & 74.93 \\
\textbf{ViT}~\cite{vit} & \textbf{43.97} & 47.36 & {\ul 86.03} & \textbf{70.57} & 73.49 & 50.88 & 59.46 & 77.89 & 73.50 & 79.15 & 36.64 & 44.90 & {\ul 66.57} & 66.78 & 71.99 \\
\textbf{DlinkViT}~\cite{my_rs} & 40.36 & 41.35 & \textbf{94.39} & 68.75 & 70.62 & 51.46 & 61.85 & 75.40 & 73.76 & 80.23 & {\ul 42.05} & {\ul 54.28} & 65.11 & {\ul 69.54} & {\ul 76.55} \\
\bottomrule
\end{tabular}%
}
\end{table*}

\begin{table*}[!t]
\caption{Quantitative results of marine fog forecasting baselines across four sub-regions (N.S., M.W., M.C and Na.) collected from the MeteoSat-11 satellite data, where the temporal length is set as $T=T'=4$.}
\label{tab:meteo_prediction}
\resizebox{\textwidth}{!}{%
\begin{tabular}{l|cccc|cccc|cccc|cccc}
\toprule
\multicolumn{1}{l|}{} & \multicolumn{4}{c|}{\textbf{N.S. sub-dataset from MeteoSat}} & \multicolumn{4}{c|}{\textbf{M.W. sub-dataset from MeteoSat}} & \multicolumn{4}{c|}{\textbf{M.C. sub-dataset from MeteoSat}} & \multicolumn{4}{c}{\textbf{Na. sub-dataset from MeteoSat}} \\
\multicolumn{1}{l|}{} & \textbf{MSE}$\downarrow$  & \textbf{MAE}$\downarrow$  & \textbf{SSIM}$\uparrow$  & \textbf{PSNR}$\uparrow$  & \textbf{MSE}$\downarrow$  & \textbf{MAE}$\downarrow$  & \textbf{SSIM}$\uparrow$  & \textbf{PSNR}$\uparrow$  & \textbf{MSE}$\downarrow$  & \textbf{MAE}$\downarrow$  & \textbf{SSIM}$\uparrow$  & \textbf{PSNR}$\uparrow$  & \textbf{MSE}$\downarrow$  & \textbf{MAE}$\downarrow$  & \textbf{SSIM}$\uparrow$  & \textbf{PSNR}$\uparrow$  \\

\midrule
\textbf{ConvLSTM}~\cite{convlstm} & 3315.46 & 17461.5 & 0.5903 & 18.07 & 2846.07 & 15810.63 & 0.5971 & 18.86 & 3638.39 & 19321.27 & 0.5016 & 17.68 & 3335.58 & 17082.83 & 0.5433 & 18.10 \\

\textbf{PredRNN}~\cite{predrnn} & \textbf{2533.55} & \textbf{14782.60} & \textbf{0.6644} & \textbf{19.43} & \textbf{2288.34} & \textbf{13382.22} & \textbf{0.6823} & {\ul 20.04} & \textbf{2864.02} & \textbf{16520.61} & \textbf{0.5888} & \textbf{18.84} & \textbf{2367.24} & \textbf{13645.51} & \textbf{0.6886} & \textbf{19.78} \\

\textbf{MIM}~\cite{mim} & {\ul 2838.82} & {\ul 15639.49} & {\ul 0.642} & {\ul 18.93} & 2727.02 & {\ul 14554.47} & {\ul 0.6530} & 19.27 & 3316.86 & {\ul 17849.62} & {\ul 0.5574} & {\ul 18.19} & {\ul 2807.68} & {\ul 14833.54} & {\ul 0.6405} & {\ul 19.02} \\

\textbf{PhyDNet}~\cite{phydnet} & 3478.27 & 17999.96 & 0.5904 & 17.94 & 2905.78 & 16085.54 & 0.6070 & \textbf{20.09} & 3600.41 & 19127.74 & 0.5245 & 17.76 & 3446.57 & 17134.09 & 0.5632 & 18.16 \\

\textbf{SimVP-v2}~\cite{simvpv2} & 3636.89 & 18112.73 & 0.5988 & 17.81 & 3221.94 & 16328.20 & 0.6206 & 18.52 & 3770.31 & 19403.51 & 0.5287 & 17.60 & 3371.52 & 16625.63 & 0.5952 & 18.24 \\

\textbf{Uniformer}~\cite{uniformer} & 3217.57 & 16676.74 & 0.6284 & 18.34 & 2896.60 & 15239.58 & 0.6435 & 19.02 & 3455.50 & 18308.89 & 0.5489 & 17.97 & 3017.33 & 15510.92 & 0.6228 & 18.70 \\

\textbf{VAN}~\cite{van} & 3575.43 & 17868.76 & 0.6029 & 17.85 & 3077.33 & 15894.65 & 0.6258 & 18.69 & 3702.35 & 19187.23 & 0.5303 & 17.66 & 3359.42 & 16457.82 & 0.5945 & 18.22 \\

\textbf{TAU}~\cite{tau} & 3086.77 & 16541.97 & 0.6279 & 18.50 & {\ul 2694.88} & 14798.46 & 0.6497 & 19.27 & {\ul 3297.54} & 17992.21 & 0.5533 & 18.15 & 2829.41 & 15084.32 & 0.6273 & 18.94 \\
\bottomrule
\end{tabular}%
}
\end{table*}

\subsection{Assessment of Regional Generalization}
We present quantitative results for detection baseline models across diverse marine fog regions using data from the Baja California (B.C.), California Current (C.C.), and Gulf of Alaska (G.A.) regions, with evaluation metrics detailed in Table~\ref{tab:XJS_exps}. Significant performance variations are observed across these sub-regions, demonstrating the effectiveness of MFogHub in assessing the regional generalization of models. Specifically, DlinkViT~\cite{my_rs} and ABCNet~\cite{abcnet} show relatively high performance in both the C.C. and G.A. regions, with C.C. being easier and G.A. more challenging. However, both models exhibit average performance in the B.C. region. In contrast, ViT-based models~\cite{vit,my_rs} perform exceptionally well in both B.C. and G.A., but their performance is weaker on the simpler C.C. dataset. Even Deeplabv3~\cite{deeplabv3+}, which performs the worst in B.C., achieves the highest precision in C.C. These results underscore the complexity of generalization. Furthermore, within the same region, we observe considerable variation in performance across different evaluation metrics, suggesting that each model’s strengths and weaknesses are metric-dependent.

Using sub-datasets from the MeteoSat satellite, specifically the North Sea (N.S.), Mediterranean West (M.W.), Mediterranean Center (M.C.), and Namibia (Na.), we obtain quantitative results for eight marine fog forecasting baseline models, as shown in Table~\ref{tab:meteo_prediction}. Compared to detection models, forecasting models exhibit stronger consistency across multiple regions. PredRNN~\cite{predrnn} consistently performs well across all four sub-regions, particularly excelling in MAE and SSIM, with MIM~\cite{mim} following closely behind. While general forecasting models demonstrate strong regional generalization ability, MFogHub reveals some regional sensitivities in certain models. For example, PhyDNet~\cite{phydnet} achieves the highest PSNR in the M.W. region, outperforming PredRNN, but performs the worst in the Na. region. Similarly, TAU~\cite{tau} achieves second place in MSE, outperforming MIM~\cite{mim} in the M.W. and M.C. regions, but lags significantly behind MIM~\cite{mim} in the N.S. region. These results highlight the capability of MFogHub to assess the regional generalization of forecasting models.

\begin{table}[!t]
\caption{Performance comparison of various marine fog detection baselines using Himawari-8 and Fengyun-4 satellites in the Yellow and Bohai Sea regions.}
\label{tab:H8_FY4A}
\resizebox{\linewidth}{!}{%
\begin{tabular}{l|ccccc|ccccc}
\toprule
\multicolumn{1}{l|}{\textbf{}} & \multicolumn{5}{c|}{\textbf{YB sub-dataset from H8/9}} & \multicolumn{5}{c}{\textbf{YB sub-dataset from FY4A}} \\
\multicolumn{1}{l|}{\textbf{}} & \textbf{CSI} $\uparrow$ & \textbf{Rec} $\uparrow$ & \textbf{Pre} $\uparrow$ & \textbf{mIoU} $\uparrow$ & \textbf{mAcc} $\uparrow$ & \textbf{CSI} $\uparrow$ & \textbf{Rec} $\uparrow$ & \textbf{Pre} $\uparrow$ & \textbf{mIoU} $\uparrow$ & \textbf{mAcc} $\uparrow$ \\

\midrule
\textbf{Deeplabv3p}~\cite{deeplabv3+} & 51.83 & 68.61 & 67.97 & 74.73 & 83.70 & 44.07 & 66.88 & 56.37 & 70.48 & 82.48 \\
\textbf{UNet}~\cite{unet} & 54.48 & 70.60 & 70.47 & 76.15 & 84.74 & 45.77 & 73.42 & 54.87 & 71.29 & 85.59 \\
\textbf{Unet++}~\cite{unetpp} & 55.71 & \textbf{73.32} & 69.94 & 76.78 & \textbf{86.07} & 45.79 & 73.38 & 54.90 & 71.30 & 85.58 \\
\textbf{ABCNet}~\cite{abcnet} & 57.36 & 70.20 & {\ul 75.99} & {\ul 77.71} & 84.68 & 46.46 & 73.44 & 55.85 & 71.68 & 87.91 \\
\textbf{BANet}~\cite{banet} & 55.11 & 67.55 & 75.02 & 76.54 & 83.35 & 46.67 & 58.72 & \textbf{69.46} & 72.11 & 78.88 \\
\textbf{Unetformer}~\cite{unetformer} & 54.63 & 68.91 & 72.50 & 76.26 & 83.97 & 48.66 & 78.03 & 56.66 & 72.93 & 87.91 \\
\textbf{ViT}~\cite{vit} & {\ul 57.03} & {\ul 72.51} & 72.62 & 77.50 & {\ul 85.75} & {\ul 52.77} & \textbf{79.38} & 61.15 & {\ul 75.08} & \textbf{88.76} \\
\textbf{DlinkViT}~\cite{my_rs} & \textbf{60.25} & 71.92 & \textbf{79.19} & \textbf{79.24} & 85.60 & \textbf{56.93} & {\ul 78.30} & {\ul 67.59} & \textbf{77.38} & {\ul 88.46} \\
\bottomrule
\end{tabular}%
}
\end{table}

\begin{table}[!t]
\caption{Performance comparison of various marine fog forecasting baselines using Himawari-8 and Fengyun-4 satellites in the Yellow and Bohai Sea region.}
\label{tab:h8_fy4a_prediction}
\resizebox{\columnwidth}{!}{%
\begin{tabular}{l|cccc|cccc}
\toprule
\multicolumn{1}{l|}{} & \multicolumn{4}{c|}{\textbf{YB sub-dataset from H8/9}} & \multicolumn{4}{c}{\textbf{YB sub-dataset from FY4A}} \\
\multicolumn{1}{l|}{} & \textbf{MSE} $\downarrow$ & \textbf{MAE} $\downarrow$ & \textbf{SSIM} $\uparrow$ & \textbf{PSNR} $\uparrow$ & \textbf{MSE} $\downarrow$ & \textbf{MAE} $\downarrow$ & \textbf{SSIM} $\uparrow$ & \textbf{PSNR} $\uparrow$ \\

\midrule
\textbf{ConvLSTM}~\cite{convlstm} & 697.12 & 7540.91 & 0.7918 & 25.31 & 1413.10 & 11077.97 & 0.6992 & 21.99 \\
\textbf{PredRNN}~\cite{predrnn} & \textbf{636.37} & \textbf{7067.29} & \textbf{0.8090} & \textbf{25.77} & 1221.37 & 10203.55 & 0.7271 & 22.63 \\
\textbf{MIM}~\cite{mim} & 668.42 & {\ul 7325.09} & {\ul 0.8002} & 25.52 & {\ul 849.78 } &{\ul 8656.75 } & \textbf{0.7538} & {\ul 24.03 }\\
\textbf{PhyDNet}~\cite{phydnet} & 803.84 & 8447.91 & 0.7770 & 24.58 & 1550.10 & 11691.48 & 0.6803 & 21.61 \\
\textbf{SimVP-v2}~\cite{phydnet} & 726.83 & 7644.08 & 0.7913 & 25.18 & \textbf{832.23} & \textbf{8514.33} & {\ul 0.7416 } & \textbf{24.07} \\
\textbf{Uniformer}~\cite{uniformer} & 679.37 & 7435.03 & 0.7988 & 25.42 & 1388.98 & 10988.48 & 0.6957 & 22.04 \\
\textbf{VAN}~\cite{van} & 728.04 & 7635.44 & 0.7908 & 25.18 & 1177.27 & 10355.11 & 0.6943 & 22.58 \\
\textbf{TAU}~\cite{tau} & {\ul 660.41} & {\ul 7273.11} & 0.7960 & {\ul 25.54} & 969.99 & 9157.88 & 0.7276 & 23.46 \\
\bottomrule
\end{tabular}%
}
\end{table}

\begin{figure*}[!t]
  \centering
  \includegraphics[width=\textwidth]{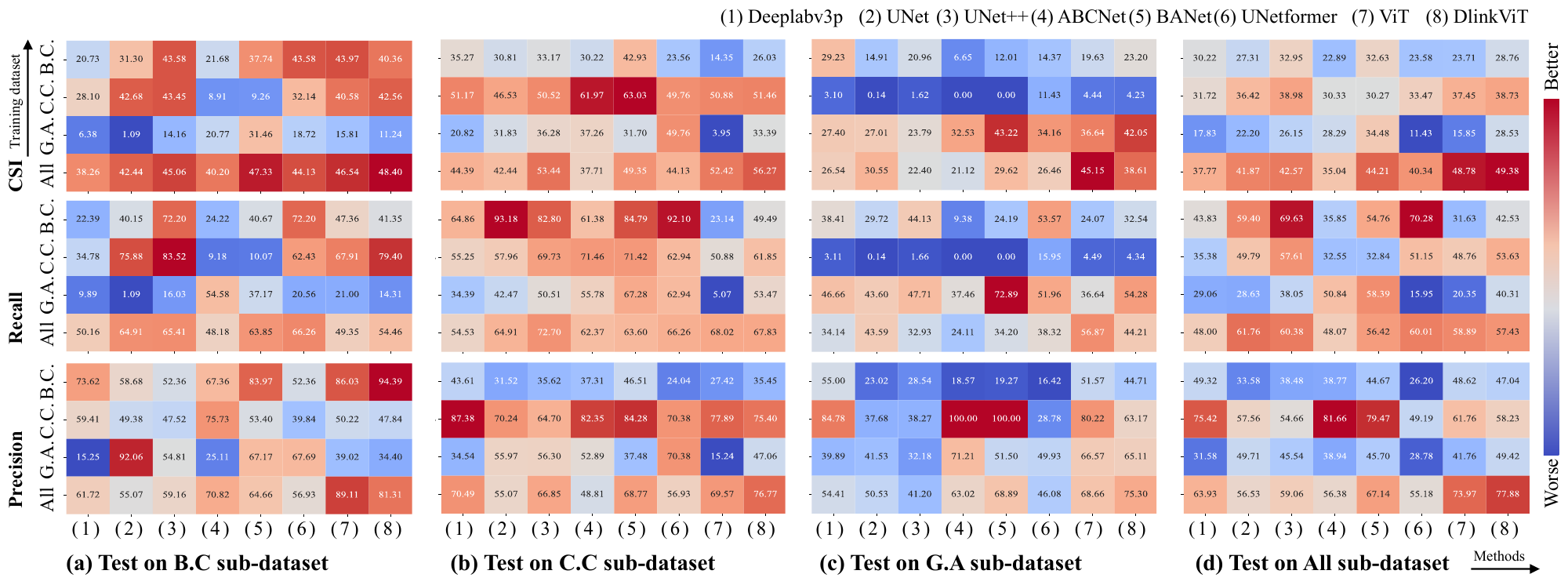}
  \caption{Multi-region evaluation of CSI, Recall and Precision metrics for eight baseline models in marine fog detection, on B.C., C.C., G.A. sub-dataset and all of them, using GOES satellite data. (a)-(d) represents testing on which sub-dataset. Each sub-figure shows rows representing training regions and columns representing model types.}
  \label{fig:heatmap_detection}
\end{figure*}

\begin{figure*}[!t]
  \centering
  \includegraphics[width=\textwidth]{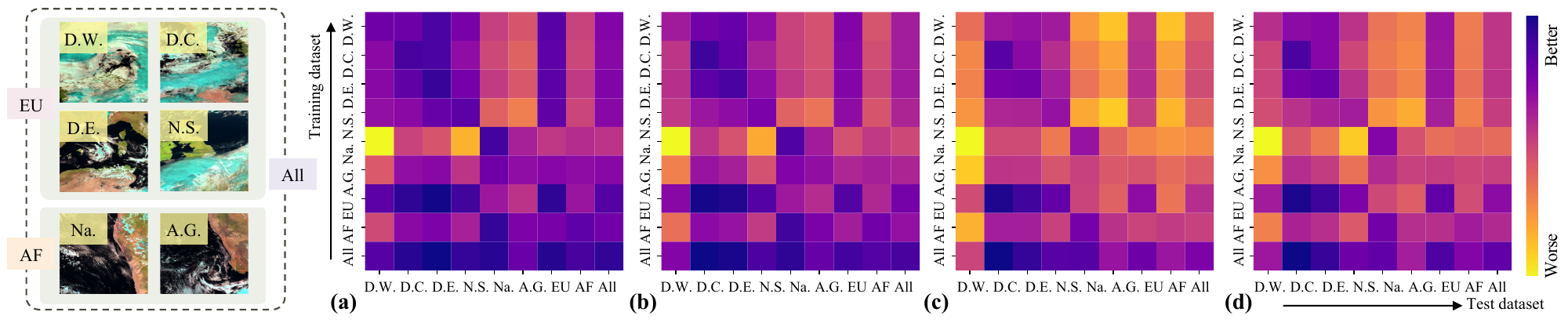}
  \caption{Performance evaluation heatmaps for forecasting using TAU~\cite{tau} method on different training and test datasets across (a) MSE, (b) MAE, (c) SSIM and (d) PSNR metrics, with single-region (D.W., D.C., D.E., N.S., Na. and A.G.), and multi-regional sub-dataset (European (EU), African (AF) and All).}
  \label{fig:heatmap_prediction}
\end{figure*}

\subsection{Assessment of satellite generalization}
\label{sec:Assessment of satellite generalization}
We use FY4A and H8 satellite data from 40 marine fog events in the Yellow and Bohai Seas (Y.B.) to evaluate satellite generalization. The results for detection and forecasting baseline models are presented in Table~\ref{tab:H8_FY4A} and Table~\ref{tab:h8_fy4a_prediction}, respectively. For detection models, high performance is concentrated among a few models. DlinkViT~\cite{my_rs} demonstrates strong cross-satellite generalization, achieving the best overall performance on both satellites. In contrast, models such as ViT~\cite{vit}, Unet++~\cite{unetpp}, ABCNet~\cite{abcnet}, and BANet~\cite{banet} show greater sensitivity to satellite-specific variations, exhibiting performance fluctuations across the two datasets. For forecasting models, PredRNN~\cite{predrnn} does not exhibit the same dominant performance as observed in the regional generalization experiments. The ranking of models differs notably between the two satellites, reflecting the influence of satellite data characteristics on performance. Notably, on the FY4A satellite, SimVP-v2~\cite{simvpv2} outperforms TAU~\cite{tau}, MIM~\cite{mim}, and PredRNN~\cite{predrnn}, securing first place. These results highlight MFogHub's effectiveness in assessing satellite generalization across different models. Furthermore, the H8/9 satellite data consistently yields better results, attributed to its additional spectral bands, reinforcing the value of multi-spectral data in improving model performance.

\subsection{Multi-regional research}
\label{sec:4.4}
Multi-region data encompass greater diversity, offering potential benefits for marine fog detection and forecasting tasks. Due to variations in the number of channels across different satellites, this study focuses on multi-region data from a single satellite. We evaluate multiple baseline models using both single-region and multi-region datasets for both tasks to assess the impact of multi-region data.

For the marine fog detection task, we train eight baseline models on GOES satellite data from individual regions (B.C., C.C., G.A.), as well as from their combined datasets. The models are evaluated using CSI, recall, and precision metrics, with the multi-region results visualized in Fig.~\ref{fig:heatmap_detection}. The results show that, when focusing on single regions, models trained and tested on the same region perform best, indicating that training on region-specific data enables better capture of local features. Cross-region evaluations highlight a domain gap, leading to performance drops. However, models trained on multi-region data consistently outperform those trained on single regions, particularly for region B.C. This suggests that incorporating additional regions into the training process enhances model generalization and stability. Notably, among the baseline models, simpler architectures and ViT-based models show the most promise for multi-region training.

For the forecasting task, we extended our datasets by combining six individual regions from Meteo satellite data into larger European, African, and all-mixed-region datasets, resulting in nine training-test sets. The results for the TAU method~\cite{tau}, evaluated using MSE, MAE, SSIM, and PSNR metrics, are visualized in Fig.~\ref{fig:heatmap_prediction}. Similar patterns emerge across these four metrics: models trained on closely related multi-region data (e.g., combined European regions) outperform those trained on single regions. This suggests that spatially proximate data are more valuable for improving prediction performance in single-region tasks.

In summary, the results from both detection and forecasting tasks demonstrate the significant benefits of incorporating multi-region data for model generalization. This highlights the potential of MFogHub to serve as a valuable datasets for developing robust marine fog models that can generalize across various regions and satellite platforms.

\subsection{Discussions}
\label{sec:4.5}

\begin{table}[!t]
\caption{Performance comparison of marine fog detection task between true-color and natural-color images.}
\label{tab:band_selection}
\resizebox{\columnwidth}{!}{%
\begin{tabular}{ll|cccc|cccc}
\toprule
\multicolumn{1}{l}{} & \multicolumn{1}{l|}{} & \multicolumn{4}{c|}{\textbf{True color images}} & \multicolumn{4}{c}{\textbf{Natural color images}} \\
\multicolumn{1}{l}{} & \textbf{} & \textbf{CSI} $\uparrow$ & \textbf{Recall} $\uparrow$ & \textbf{mIoU} $\uparrow$ & \textbf{mAcc} $\uparrow$ & \textbf{CSI} $\uparrow$ & \textbf{Recall} $\uparrow$ & \textbf{mIoU} $\uparrow$ & \textbf{mAcc} $\uparrow$ \\

\midrule
\multirow{4}{*}{\textbf{H8/9}} & \textbf{UNet}~\cite{unet} & 41.95 & 69.46 & 59.12 & 78.79 & \cellcolor[HTML]{F7EDF3}51.22 & \cellcolor[HTML]{F7EDF3}74.36 & \cellcolor[HTML]{F7EDF3}70.67 & \cellcolor[HTML]{F7EDF3}84.62 \\

 & \textbf{Unetformer}~\cite{unetformer} & \cellcolor[HTML]{F7EDF3}\textbf{53.08} & \cellcolor[HTML]{F7EDF3}\textbf{75.44} & 67.04 & 83.03 & 52.40 & 74.97 & \cellcolor[HTML]{F7EDF3}72.80 & \cellcolor[HTML]{F7EDF3}85.67 \\

 & \textbf{ViT}~\cite{vit} & 51.98 & 74.84 & \textbf{67.46} & \textbf{83.17} & \cellcolor[HTML]{F7EDF3}57.39 & \cellcolor[HTML]{F7EDF3}77.70 & \cellcolor[HTML]{F7EDF3}72.32 & \cellcolor[HTML]{F7EDF3}85.67 \\

 & \textbf{DlinkViT}~\cite{my_rs} & 51.27 & 74.49 & 65.02 & 82.01 & \cellcolor[HTML]{F7EDF3}\textbf{58.10} & \cellcolor[HTML]{F7EDF3}\textbf{78.01} & \cellcolor[HTML]{F7EDF3}\textbf{77.55} & \cellcolor[HTML]{F7EDF3}\textbf{88.15} \\

\midrule
\multirow{4}{*}{\textbf{FY4A}} & \textbf{UNet}~\cite{unet} & 37.55 & 67.16 & 53.46 & 75.94 & \cellcolor[HTML]{F7EDF3}50.81 & \cellcolor[HTML]{F7EDF3}74.25 & \cellcolor[HTML]{F7EDF3}65.28 & \cellcolor[HTML]{F7EDF3}82.11 \\

 & \textbf{Unetformer}~\cite{unetformer} & \textbf{51.38} & \textbf{74.59} & \textbf{63.41} & \textbf{81.27} & \cellcolor[HTML]{F7EDF3}54.15 & \cellcolor[HTML]{F7EDF3}76.04 & \cellcolor[HTML]{F7EDF3}67.21 & \cellcolor[HTML]{F7EDF3}83.16 \\

 & \textbf{ViT}~\cite{vit} & 49.40 & 73.52 & 63.19 & 81.08 & \cellcolor[HTML]{F7EDF3}53.17 & \cellcolor[HTML]{F7EDF3}75.44 & \cellcolor[HTML]{F7EDF3}\textbf{71.12} & \cellcolor[HTML]{F7EDF3}\textbf{84.94} \\

 & \textbf{DlinkViT}~\cite{my_rs} & 50.06 & 73.93 & 60.37 & 79.81 & \cellcolor[HTML]{F7EDF3}\textbf{54.84} & \cellcolor[HTML]{F7EDF3}\textbf{76.35} & \cellcolor[HTML]{F7EDF3}70.63 & \cellcolor[HTML]{F7EDF3}84.78 \\
\bottomrule
\end{tabular}%
}
\end{table}

\paragraph{Impact of different spectral band combinations on model performance.} Building on our analysis of fog sensitivity for each band in Sec.~\ref{sec:Statistics_Analysis}, we further investigate the role of various spectral bands in marine fog detection by constructing true-color (using the first three channels) and natural-color images (composed of the 0.64$\mu$m, 0.86$\mu$m, and 3.9$\mu$m bands) through different band combinations. Using satellite data from the H8/9 and FY4A satellites as examples, the image synthesis method and data are illustrated in Fig.~\ref{fig:analysis_01} (c). Four baseline models~\cite{unet, unetformer, vit, my_rs} are selected to evaluate how different band combinations impact marine fog detection within the same model architecture. The results indicate that natural-color images outperform true-color images, as shown by the pink shading in the table. This suggests that bands with greater pixel distribution differences between fog and non-fog areas are more valuable for fog detection, particularly when working with limited-channel configurations. However, as demonstrated in Table~\ref{tab:H8_FY4A}, models utilizing all available bands outperform those using any three-band combination, implying that the full-band data may contain additional inter-relationships that are beneficial for marine fog detection.

\begin{figure}[!t]
  \centering
  \includegraphics[width=\linewidth]{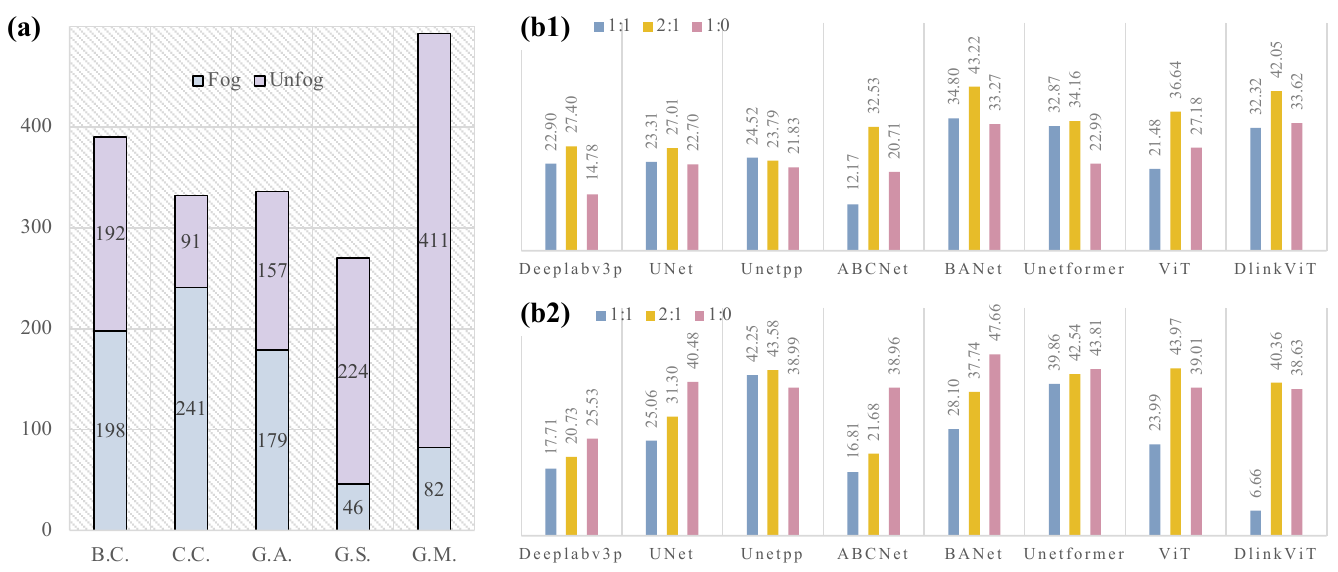}
  \caption{(a) Proportion of positive and negative samples across different regions from GOES satellite. (b1) and (b2) comparison of model performance under varying pos-neg ratios (1:1, 2:1, 1:0) in B.C. and G.A. sub-dataset with CSI metric, respectively. }
  \label{fig:exps_posneg}
\end{figure}

\paragraph{Impact of positive-to-negative sample ratios.} Marine fog is a seasonal phenomenon with irregular occurrences, leading to significant variation in the positive-to-negative sample ratios across different regions during dataset collection. For example, in the five offshore regions captured by the GOES satellite (as shown in Fig.~\ref{fig:exps_posneg} (a)), the Gulf of Mexico (G.M.) sub-dataset has the lowest proportion of positive samples, while the B.C. and G.A. sub-datasets maintain a 1:1 ratio. Based on this, we design three training settings for the B.C. and G.A. sub-datasets, with positive-to-negative sample ratios of 1:1, 2:1, and 1:0, while keeping the number of positive samples constant across all training sets. We then evaluate model performance on a common validation set. The results, specifically in terms of the CSI metric (see Fig.~\ref{fig:exps_posneg} (b1) and (b2)), indicate that different negative sample ratios affect model training to varying degrees; additional metrics, such as recall and precision, are provided in the supplementary materials. Notably, Unet++~\cite{unetpp} demonstrated stable performance across different negative sample ratios, indicating strong robustness against negative sample disturbances. Based on the results from both regions, we recommend a positive-to-negative sample ratio of 2:1 for future dataset construction. This ratio strikes a favorable balance between missed detections and false alarms, making it particularly suitable for marine fog monitoring tasks.

\section{Conclusions}
In this work, we introduce MFogHub, the most comprehensive, publicly available dataset for global, multi-regional, and multi-satellite marine fog detection and forecasting to date. MFogHub comprises 68,000 data samples from 15 coastal regions prone to marine fog, collected from six geostationary satellites. The data is organized in a ``cube-stream" format, consisting of 21 streams in total. Extensive experiments have demonstrated that MFogHub can assess the impact of regional and satellite differences on model performance, while also supports leveraging regional diversity to enhance the model's representation capabilities. We hope MFogHub will bridge natural sciences and AI, driving both technological and practical advancements.

\section*{Acknowledgement}

We sincerely thank our colleagues and collaborators for their invaluable support and insightful feedback throughout this research. We also express our gratitude to institutions and organizations, such as the China National Meteorological Centre (NMC), for providing the necessary resources and facilities. This research is supported by National Natural Science Foundation of China (NSFC) under Grant U24B20177 and Hebei Natural Science Foundation Project No. F2024502017. Additionally, this research is supported by Super Computing Platform of Beijing University of Posts and Telecommunications.

{
    \small
    \bibliographystyle{ieeenat_fullname}
    \bibliography{main}
}



\end{document}